\documentclass[letterpaper]{article} 
\usepackage{aaai2027}
\usepackage[hyphens]{url}  
\usepackage{graphicx} 
\usepackage{natbib}  
\usepackage{caption} 
\usepackage{algorithm}
\usepackage{algorithmic}

\nocopyright

\usepackage{newfloat}
\usepackage{listings}
\DeclareCaptionStyle{ruled}{labelfont=normalfont,labelsep=colon,strut=off} 
\floatstyle{ruled}
\newfloat{listing}{tb}{lst}{}
\floatname{listing}{Listing}

\usepackage{booktabs}

\usepackage{multirow}
\usepackage{amsmath}
\usepackage{amssymb}

\title{From Visual Matching to Textual Conditioning:\\Cross-Domain Few-Shot Segmentation via Reference Inversion}
\author{
   Jiahao Nie\textsuperscript{\rm 1}\equalcontrib \ \ Yun Xing\textsuperscript{\rm 1}\equalcontrib \ \ \ Wenbin An\textsuperscript{\rm 2} \ \ Qixin Zhang\textsuperscript{\rm 1}\\
   Yap-Peng Tan\textsuperscript{\rm 3,\rm 1} \ \ \ Alex C. Kot\textsuperscript{\rm 4} \ \ \ Shijian Lu\textsuperscript{\rm 1}\corresponding
}
\affiliations{
    \textsuperscript{\rm 1} Nanyang Technological University \ \ \ \textsuperscript{\rm 2} Xi'an Jiaotong University\\
    \textsuperscript{\rm 3} VinUniversity \ \ \ \ \textsuperscript{\rm 4} Shenzhen MSU-BIT University\\
    jiahao007@e.ntu.edu.sg \ \ \ \ \ \ shijian.lu@ntu.edu.sg
}

\begin{document}

\maketitle

\begin{abstract}
Cross-Domain Few-Shot Segmentation (CD-FSS) aims to segment objects from unseen domains using only a few annotated references. Recent Segment Anything Model (SAM)-based methods have advanced CD-FSS, yet their reference-target visual matching is highly sensitive to target appearance variations. SAM3 introduces text-conditioned segmentation, enabling natural-language descriptions to guide segmentation across target images, but directly using the category name as the condition still yield unstable performance under severe domain shifts. We propose Reference Inversion for SAM3 (RI-SAM3) that leverages SAM3’s large-scale vision-language pre-training to learn suitable textual tokens for each target category. RI-SAM3 comprises three core designs: \textit{(i)} Bi-level Reference Inversion that learns complementary tokens before and after the frozen text encoder: the former remain anchored in SAM3’s text space while the latter represent the category more flexibly; \textit{(ii)} Layer-adaptive Textual Conditioning combines these tokens differently as object queries evolve across decoder layers; and \textit{(iii)} Presence-aware Confidence Supervision that jointly supervises query confidence and target presence for reliable predictions. With SAM3 frozen, RI-SAM3 optimizes around 12K parameters only per category and it efficiently segments each target in a single forward pass with reusable textual conditioning without per-sample reference-target visual matching. Extensive experiments over multiple widely adopted benchmarks show that RI-SAM3 achieves superior performance consistently in CD-FSS tasks, together with strong robustness, efficiency, and generalization.

\end{abstract}
\section{Introduction}

Recent advances in semantic segmentation~\cite{cheng2022masked,kirillov2023segment,wang2024image} have achieved remarkable performance, yet most existing methods still rely heavily on large-scale annotated data. Such annotations are costly or impractical to obtain in data-scarce domains, including medical~\cite{codella2019skin}, satellite~\cite{demir2018deepglobe}, and underwater imagery~\cite{islam2020semantic}. Cross-Domain Few-Shot Segmentation (CD-FSS)~\cite{lei2022cross} alleviates this burden by segmenting target categories from previously unseen domains using only a few annotated reference exemplars of the same categories. Conventional CD-FSS methods~\cite{su2024domain,nie2026cross,fan2024adapting,tong2024lightweight} train a few-shot segmentation model on a large-scale source domain and then adapt it to each target domain. Although effective, this paradigm requires heavy source-domain training and target-domain adaptation, limiting its flexibility and efficiency.

\begin{figure*}[t]
    \centering
    \includegraphics[width=\linewidth]{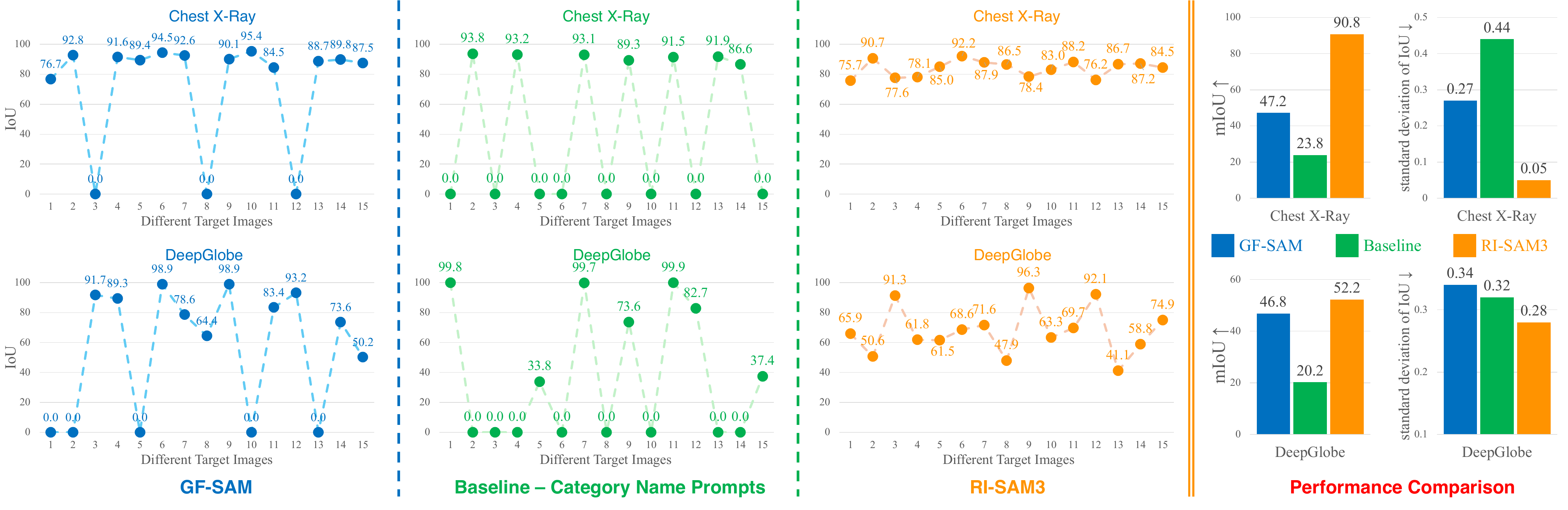}
    \caption{Motivation for Reference Inversion for SAM3 (RI-SAM3). \textbf{Left}: Using the same reference image, GF-SAM~\cite{zhang2024bridge} exhibits substantial IoU variations across different target images, while manually specified text prompts produce low and unstable results under severe domain shifts. Our proposed RI-SAM3 instead inverts annotated references into a reusable textual condition for inference targets. \textbf{Right}: RI-SAM3 achieves higher mIoU ($\uparrow$) and a lower standard deviation of per-image IoU ($\downarrow$) than GF-SAM and the category-name prompt baseline, demonstrating improved accuracy and stability.}
    \label{fig:motivation}
\end{figure*}

To reduce this dependence, recent CD-FSS methods~\cite{liu2023matcher,zhang2024bridge,xu2025unlocking,nie2026boosting} match reference and target images using the transferable dense features of DINOv2~\cite{oquab2023dinov2}, and convert the resulting correspondences into geometric prompts that guide SAM~\cite{kirillov2023segment} to segment the target images. Although reducing training cost, this paradigm exhibits weak generalization across target images of the same category. As illustrated in Fig.~\ref{fig:motivation}~(Left), GF-SAM~\cite{zhang2024bridge} exhibits substantial variations in prediction quality across target images, even when using the same reference image. This variation suggests that visual matching depends on the appearance similarity between each reference and target image, rather than representing the target category consistently across images. Moreover, the reference image remains necessary during inference, as visual matching must be conducted for every target image. These limitations motivate us to seek a category representation that can be reused across target images.

Natural language offers such a representation by describing categories without relying on visual references~\cite{radford2021learning}. SAM3~\cite{carion2025sam3} further supports text-conditioned segmentation, allowing these language descriptions to guide target segmentation. A straightforward attempt is to use category names as SAM3's textual prompts. However, as shown in Fig.~\ref{fig:motivation}~(Left)\footnote{For Chest X-Ray, we use ``lung'' as the text prompt. For DeepGlobe, we use ``forest land'' for the forest category. Additional quantitative comparisons among manually specified prompts and qualitative examples are provided in the appendix.}, this strategy yields low average performance and substantial variations across targets of the same category. These results indicate that category-name prompts still yield poor generalization across targets under severe domain shifts. Nevertheless, SAM3's large-scale vision-language pre-training provides a text space aligned with visual categories, motivating us to seek resuable tokens that represent each target category within this text space.

Textual inversion for personalized visual generation~\cite{gal2022image} offers a way to obtain such a representation by inverting new textual tokens from a few references. Motivated by its success, we propose Reference Inversion for SAM3 (RI-SAM3), which converts annotated references into reusable soft textual tokens. The references are used only during inversion and discarded during inference, replacing per-sample reference-target visual matching with reusable textual conditioning. To make the inversion effective under substantial domain shifts, RI-SAM3 incorporates three designs. First, Bi-level Reference Inversion learns complementary tokens before and after the frozen text encoder: the former remain anchored in SAM3's text space, while the latter represent the target category more flexibly. Second, as object queries evolve through the decoder, Layer-adaptive Textual Conditioning combines these tokens differently across layers. Third, Presence-aware Confidence Supervision jointly supervises query confidence and target presence to improve prediction reliability. With the SAM3 backbone frozen, RI-SAM3 optimizes only approximately 12K parameters per category. The inverted tokens then condition SAM3 to segment each target image in a single forward pass, without references or per-sample visual matching during inference. As shown in Fig.~\ref{fig:motivation}~(Right), RI-SAM3 achieves higher mIoU and lower standard deviation across target images than GF-SAM and the category-name baseline, demonstrating improved accuracy and stability. Extensive experiments across widely adopted benchmarks further demonstrate substantial improvements over existing CD-FSS methods, together with strong robustness, efficiency, and generalization.

The contributions of this work can be summarized in three aspects: \textbf{\textit{First}}, we revisit reference-target visual matching in CD-FSS and identify its limited generalization across targets of the same category, as predictions are sensitive to variations in target appearance. \textbf{\textit{Second}}, we propose RI-SAM3, a lightweight Reference Inversion framework that converts annotated references into reusable soft textual tokens through three complementary designs. With SAM3 frozen, RI-SAM3 optimizes only approximately 12K parameters per category and conditions each target prediction in a single forward pass without references or visual matching. \textbf{\textit{Third}}, extensive experiments show that RI-SAM3 achieves strong performance on CD-FSS benchmarks while exhibiting robustness to different reference images, high computational efficiency, and generalization to related few-shot tasks.
\section{Related Work}\label{sec:related_work}

\subsection{Cross-Domain Few-Shot Segmentation}

Cross-Domain Few-Shot Segmentation (CD-FSS) aims to segment target objects conditioned on a few annotated references under significant domain shifts~\cite{lu2021simpler,wang2022remember,tong2025self,wu2024task,chen2024pixel,zhu2025cross,tian2025high,nie2026cross,cong2026divide}. Early methods~\cite{lei2022cross,fu2024cross} usually follow a two-stage paradigm, training on a large-scale source domain and then adapting to each target domain through dynamic refinement~\cite{fan2023darnet}, knowledge transfer~\cite{huang2023restnet}, frequency disentanglement~\cite{tong2024lightweight}, or lightweight adaptation~\cite{fan2024adapting,su2024domain}. With the advent of vision foundation models~\cite{kirillov2023segment,oquab2023dinov2,rombach2022high,ravi2024sam2}, recent studies~\cite{liu2023matcher,zhang2024bridge,zhu2024unleashing,sun2024vrp,liu2024simple,cuttano2025sansa,peng2025sam,xu2025unlocking,nie2026boosting,ma2026selective} typically conduct reference-target visual matching and convert the correspondences into geometric prompts for SAM. However, such correspondences remain specific to individual reference-target pairs and can be sensitive to target appearance variations. SAM3~\cite{carion2025sam3} introduces text-conditioned segmentation, which can segment all target images with natural language prompts. Motivated by this capability, we replace pairwise visual matching with textual conditioning using reusable soft reference tokens to segment all target images.

\subsection{SAM Adaptation for Downstream Tasks}

Despite their strong promptable segmentation ability, SAM-family models~\cite{kirillov2023segment,ravi2024sam2} often require adaptation to specialized domains, including medical~\cite{chen2023samadapter,ma2024medsam,wu2025medicalsamadapter} and satellite imagery~\cite{li2025segearthov,ding2024samcd,liu2025pointsam}. Existing approaches  adapt SAM using large-scale domain-specific annotations~\cite{ma2024medsam}, lightweight trainable modules~\cite{chen2023samadapter,ke2023samhq}, or few-shot target-domain tuning~\cite{xiao2024cat}. These methods primarily adapt the model representation or mask decoder to a downstream domain. In contrast, we keep SAM3 frozen and adapt its input prompts by learning only reference tokens and lightweight layer-wise coefficients.

\subsection{Textual Inversion for Visual Generation}

Textual inversion~\cite{gal2022image} is a representative approach for customized content generation with pre-trained text-to-image models~\cite{podell2024sdxl}. Given a few references of a specific subject, it learns new tokens in the text space, enabling this subject to be re-generated in novel scenes. Subsequent personalization methods~\cite{ruiz2023dreambooth} have extended this idea to multiple subjects~\cite{jiang2025mc2}, video motions~\cite{huang2025videomage}, and 3D assets~\cite{raj2023dreambooth3d}. Like customized content generation, target categories under substantial domain shifts cannot be fully represented by manually specified language prompts. Inspired by the success of textual inversion, we introduce reference inversion to transform annotated references into reusable reference tokens that condition SAM3 to segment any target image of the same category.

\section{Preliminary Study}

\subsection{Problem Formulation}

Cross-Domain Few-Shot Segmentation (CD-FSS) aims to segment target objects from previously unseen domains with only a few annotated references. In the $K$-shot setting, a reference set $\mathcal{S}_c=\{(I_r^i,M_r^i)\}_{i=1}^{K}$ is provided for category $c$, where $I_r^i$ and $M_r^i$ denote the $i$-th reference image and its ground-truth mask, respectively. Given a target image $I_t$ of the same category, the goal is to predict its segmentation mask $M_t$ using the available reference set $\mathcal{S}_c$.

\noindent\textbf{Reference-Target Visual Matching.} Previous CD-FSS methods generally condition each target prediction directly on the references: $\hat{M}_t=\mathcal{G}_{\Phi}(I_t,\mathcal{S}_c),$ where $\mathcal{G}_{\Phi}$ denotes a CD-FSS model. Regardless of the underlying architecture, these methods typically match reference and target features through visual prototypes, feature correlations, or dense correspondences. Consequently, each target prediction depends directly on $\mathcal{S}_c$, requiring reference-target visual matching to be performed for every $I_t$. Therefore, such visual similarity is sensitive to appearance variations (refer to Fig.~\ref{fig:motivation}).

\noindent\textbf{Reference Inversion.} In this paper, we reformulate the role of references $\mathcal{S}_c$ in CD-FSS. Instead of performing visual matching between $\mathcal{S}_c$ and each $I_t$, we invert the references into a reusable textual condition $T_c^*$ that can be shared across target images. Let $\mathcal{F}_{\Theta}$ denote the pre-trained SAM3 with frozen parameters $\Theta$. Given $\mathcal{S}_c$, we obtain $T_c^*$ while keeping $\Theta$ frozen. After inversion, $\mathcal{S}_c$ is discarded, and target prediction is performed as $\hat{M}_t=\mathcal{F}_{\Theta}(I_t;T_c^*).$ This formulation decouples references from target inference, allowing $T_c^*$ to be reused across all target images without further access to $\mathcal{S}_c$. Importantly, this reformulation does not modify the standard CD-FSS reference-target construction; it only replaces pairwise visual matching with a reusable conditioning representation.

\subsection{Segment Anything Model 3}

\noindent\textbf{Architecture.} Segment Anything Model 3 (SAM3)~\cite{carion2025sam3} extends the Segment Anything family~\cite{kirillov2023segment,ravi2024sam2} to promptable concept segmentation, where concept prompts, such as category names, specify all matching instances to be segmented. For text-conditioned segmentation, SAM3 encodes the text prompt and input image into textual and visual tokens, respectively. Its mask decoder follows a DETR-style architecture~\cite{carion2020end}, progressively refining object-query tokens through self-attention, query-text cross-attention, and query-visual cross-attention. Each decoder layer predicts a classification score and a box offset for each object-query token, while a segmentation head decodes the final query tokens into instance masks. SAM3 further introduces a global presence token to estimate whether the prompted concept appears in the image, with its presence score modulating the query-level classification scores.

\noindent\textbf{Design Implications.} SAM3's predictions depend critically on how its textual prompt represents the target category. Category names provide only generic descriptions, motivating us to invert references into complementary word- and concept-level tokens. Moreover, visual representations at different depths encode complementary semantic information~\cite{hariharan2015hypercolumns}, while stacked decoder layers progressively refine object queries~\cite{carion2020end}. Using an identical textual condition across all decoder layers may therefore be suboptimal, motivating layer-wise combinations of the two tokens. Finally, because mask confidence depends jointly on query and presence scores, both require reliable supervision under domain shifts. These properties motivate our Bi-level Reference Inversion, Layer-adaptive Textual Conditioning, and Presence-aware Confidence Supervision.
\begin{figure*}[t]
    \centering
    \includegraphics[width=0.98\linewidth]{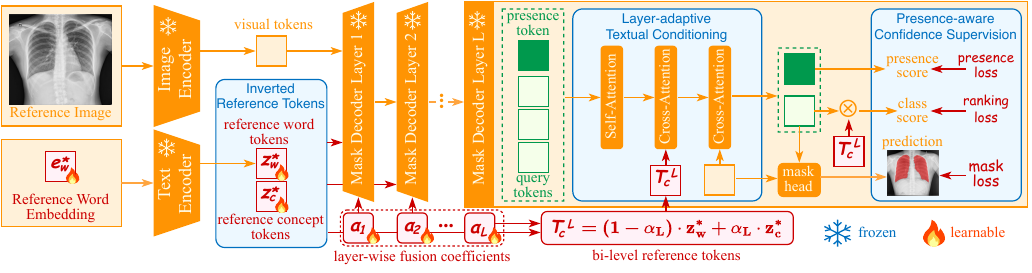}
    \caption{Overview of our proposed RI-SAM3 framework. Bi-level Reference Inversion (BRI) learns reference word tokens $z_w^*$ and reference concept tokens $z_c^*$ from annotated references, collectively termed inverted reference tokens. Layer-adaptive Textual Conditioning (LTC) combines them with layer-specific fusion coefficients $\alpha_*$ to produce the bi-level reference tokens $T_c^*$ for conditioning frozen SAM3. Presence-aware Confidence Supervision (PCS) jointly supervises query-level classification and global presence scores. Once learned, $T_c^*$ is reused for target segmentation without references or visual matching.}
    \label{fig:model}
\end{figure*}

\section{Method}

As illustrated in Fig.~\ref{fig:model}, our Reference Inversion framework transforms annotated references into a reusable textual representation while keeping SAM3 frozen. Bi-level Reference Inversion optimizes a reference word embedding $\mathbf{e}_w^*$ before the text encoder and the reference concept tokens $\mathbf{z}_c^*$ after it. Layer-adaptive Textual Conditioning combines the encoded reference word tokens $\mathbf{z}_w^*$ and $\mathbf{z}_c^*$ with layer-wise fusion coefficients $\alpha_*$ to construct the bi-level reference tokens $T_c^*$ for conditioning different decoder layers. Presence-aware Confidence Supervision further improves prediction reliability by supervising mask quality, query classification, and target presence. Once inverted, $T_c^*$ is reused across target images without further access to the annotated references.

\subsection{Bi-level Reference Inversion}

Textual inversion learns a pseudo-word embedding at the input of a frozen text encoder~\cite{gal2022image}. Passing the learned pseudo-word embedding through the text encoder anchors the resulting tokens to the pre-trained language space, but may limit its flexibility in capturing target-category appearance under severe domain shifts. In contrast, directly optimizing the tokens after the text encoder provides greater flexibility but lacks such a language anchor. We therefore perform Reference Inversion at both levels.

Let $\mathcal{E}_{\mathrm{text}}$ denote the frozen text encoder of SAM3. We introduce a learnable reference word embedding $\mathbf{e}_w$ at its input and optimize it using the annotated references. The frozen text encoder maps $\mathbf{e}_w$ to the reference word tokens:
\begin{equation}
\mathbf{z}_w^*=\mathcal{E}_{\mathrm{text}}(\mathbf{e}_w^*).
\end{equation}
Because $\mathbf{z}_w^*$ is obtained through the frozen text encoder, the resulting tokens remain anchored in SAM3's pre-trained text space. Meanwhile, we introduce reference concept tokens $\mathbf{z}_c^*$ directly after the text encoder. Without being constrained by the text encoding process, $\mathbf{z}_c^*$ can more flexibly represent the target-category concept specified by the references. They are independently optimized using the annotated references:
\begin{align}
\mathbf{e}_w^*&=\arg\min_{\mathbf{e}_w}\frac{1}{K} \sum_{i=1}^{K} \mathcal{L}_{\mathrm{ref}} ( \mathcal{F}_{\Theta} ( I_r^i; \mathcal{E}_{\mathrm{text}}(\mathbf{e}_w) ), M_r^i ),\\
\mathbf{z}_c^*&=\arg\min_{\mathbf{z}_c} \frac{1}{K} \sum_{i=1}^{K} \mathcal{L}_{\mathrm{ref}} ( \mathcal{F}_{\Theta} ( I_r^i;\mathbf{z}_c ), M_r^i),
\end{align}
where $\mathcal{L}_{\mathrm{ref}}$ represents supervision introduced below. The resulting reference word tokens $\mathbf{z}_w^*=\mathcal{E}_{\mathrm{text}}(\mathbf{e}_w^*)$ and reference concept tokens $\mathbf{z}_c^*$ are collectively referred to as the inverted reference tokens.

\begin{table*}[t]

\renewcommand\arraystretch{1.05}
\centering

\small
\setlength{\tabcolsep}{9.4pt}

\begin{tabular}{l||c||cc|cc|cc|cc}

\toprule[1pt]
\multirow{2}*{Method} & \multirow{2}*{Venue} & \multicolumn{2}{c|}{ISIC} & \multicolumn{2}{c|}{Chest X-Ray} & \multicolumn{2}{c|}{DeepGlobe} & \multicolumn{2}{c}{SUIM}\\ \cline{3-10}

& & 1-shot & 5-shot & 1-shot & 5-shot & 1-shot & 5-shot & 1-shot & 5-shot\\ \hline\hline
\multicolumn{10}{c}{\textit{\textbf{Conventional ResNet-/ViT-based Methods}}}\\ \hline

ISA & ICCV 2025 & 37.2 & 56.1 & 83.4 & 86.3 & 44.3 & 52.7 & \textbf{44.1} & --\\
DR-Adapter & CVPR 2024 & 40.8 & 48.9 & 82.4 & 82.3 & 41.3 & 50.1 & -- & --\\
PATNet & ECCV 2022 & 41.2 & 53.6 & 66.6 & 70.2 & 37.9 & 43.0 & 32.1 & 40.2\\
APM & NeurIPS 2024 & 41.7 & 51.2 & 78.3 & 82.8 & 40.9 & 44.9 & -- & --\\
ABCDFSS & CVPR 2024 & 45.7 & 53.3 & 79.8 & 81.4 & 42.6 & 49.0 & 35.1 & \textbf{41.3}\\
SDRC & ICML 2025 & 46.6 & 55.0 & 82.9 & 84.8 & 43.2 & 46.8 & -- & --\\
DFN & ICML 2025 Spotlight & 50.4 & 58.5 & 83.2 & \textbf{87.1} & 39.5 & 47.7 & -- & --\\
LoEC & CVPR 2025 & 52.9 & 62.4 & \textbf{83.9} & 84.1 & 42.1 & 51.5 & -- & --\\
IFA & CVPR 2024 & 66.3 & 69.8 & 74.0 & 74.6 & 50.6 & 58.8 & -- & --\\
DATO & CVPR 2025 & \textbf{68.8} & \textbf{70.3} & 79.6 & 81.1 & \textbf{51.1} & \textbf{59.3} & -- & --\\
\hline\hline

\multicolumn{10}{c}{\textit{\textbf{Vision Foundation Model-based Methods}}}\\ \hline
PerSAM & ICLR 2024 & 23.9 & -- & -- & -- & 31.4 & -- & -- & --\\
Matcher & ICLR 2024 & 38.6 & 35.0 & 66.8 & 66.3 & 48.1 & 50.9 & 37.9 & 41.4\\
APSeg & CVPR 2024 & 45.4 & 54.0 & 84.1 & 84.5 & 35.9 & 40.0 & -- & --\\
TAVP & arXiv 2024 & 54.9 & -- & -- & -- & 46.1 & -- & -- & --\\
CPS & arXiv 2026 & 51.3 & 56.6 & 70.0 & 70.2 & 46.0 & 49.8 & 42.8 & 48.8\\
GF-SAM & NeurIPS 2024 Spotlight & 48.7 & 55.2 & 47.6 & 47.4 & 49.5 & \textbf{57.7} & 38.3 & 42.9\\
PerSAM-F$^*$ & ICLR 2024 & 20.5 & -- & 38.7 & -- & 46.5 & -- & 10.3 & --\\
VLP-SAM$^*$ & arXiv 2025 & 34.8 & 45.9 & 54.7 & 59.1 & 48.1 & 52.4 & 16.9 & 26.9\\
INSID3 & CVPR 2026 Oral & 54.4 & 63.9 & 78.8 & 80.1 & -- & -- & 54.9 & 61.7\\
\textbf{RI-SAM3}$^*$ & -- & \textbf{64.1} & \textbf{71.4} & \textbf{90.5} & \textbf{91.0} & \textbf{51.8} & 55.8 & \textbf{63.2} & \textbf{78.1}\\

\bottomrule[1pt]
\end{tabular}

\caption{Quantitative comparison between the proposed RI-SAM3 and existing methods on four CD-FSS benchmarks in mIoU (\%). Results are taken from the original papers whenever available. Methods marked by $^*$ are averaged over five random seeds. The best result within each method group is highlighted in \textbf{bold}.}
\label{tab:cdfss}

\end{table*}

\subsection{Layer-adaptive Textual Conditioning}

SAM3 progressively refines its object-query tokens through $L$ decoder layers, each consisting of query self-attention, query-text cross-attention, and query-visual cross-attention. Since the object queries evolve across decoder layers, using the same textual condition throughout the decoder may be suboptimal. We therefore introduce Layer-adaptive Textual Conditioning, which adaptively combines the two inverted reference tokens for each decoder layer. Specifically, the bi-level reference tokens for the $l$-th layer are constructed as:
\begin{equation}
\mathbf{T}_c^l = (1-\alpha_l)\mathbf{z}_w^* + \alpha_l\mathbf{z}_c^*, \qquad \alpha_l\in[0,1],
\end{equation}
where $\alpha_l$ is the learnable fusion coefficient of the $l$-th decoder layer. Omitting residual connections, normalization, and feed-forward layers for clarity, the object-query tokens are updated as:
\begin{align}
\overline{\mathbf{Q}}_l &= \operatorname{SelfAttn}_l (\mathbf{Q}_{l-1}),\\
\widetilde{\mathbf{Q}}_l &= \operatorname{CrossAttn}^{\mathrm{text}}_l (\overline{\mathbf{Q}}_l,\mathbf{T}_c^l),\\
\mathbf{Q}_l &= \operatorname{CrossAttn}^{\mathrm{vis}}_l (\widetilde{\mathbf{Q}}_l,\mathbf{v} ),
\end{align}
where $\mathbf{v}$ denotes the visual tokens of the input image. In this way, each decoder layer receives a textual condition adapted to its current object-query tokens before interacting with the visual tokens.

With the inverted reference tokens fixed, we optimize the layer-wise fusion coefficients: $\boldsymbol{\alpha}=\{\alpha_l\}_{l=1}^{L}$ using the annotated references:
\begin{equation}
\boldsymbol{\alpha}^* = \operatorname*{arg\,min}_{\boldsymbol{\alpha}} \frac{1}{K} \sum_{i=1}^{K} \mathcal{L}_{\mathrm{ref}} ( \mathcal{F}_{\Theta} (I_r^i;T_c(\boldsymbol{\alpha})), M_r^i ),
\end{equation}
where $T_c(\boldsymbol{\alpha}) = \{\mathbf{T}_c^l\}_{l=1}^{L}$ collects the layer-specific bi-level reference tokens. Once optimized, $\boldsymbol{\alpha}^*$ remains fixed and is shared across all target images of category $c$.

\subsection{Presence-aware Confidence Supervision}

SAM3 determines the confidence of each predicted mask using both its query-level classification score and the global presence score. Under substantial domain shifts, conventional mask supervision does not directly ensure the reliability of either score. We therefore introduce Presence-aware Confidence Supervision to align query confidence with mask quality and supervise target presence. Let $\mathbf{A}_j$ and $s_j$ denote the mask and classification logits of the $j$-th object query, respectively, and let $s_p$ denote the global presence logit. We first apply standard BCE and Dice losses to the reference prediction $\hat{M}_r=\max_j\sigma(\mathbf{A}_j)$:
\begin{equation}
\mathcal{L}_{\mathrm{mask}} = \mathcal{L}_{\mathrm{BCE}}(\hat{M}_r,M_r) + \mathcal{L}_{\mathrm{Dice}}(\hat{M}_r,M_r).
\end{equation}
For each query, we use its soft mask IoU as the quality target:
\begin{equation}
q_j = \operatorname{IoU} ( \sigma(\mathbf{A}_j),M_r ), \qquad j^*=\arg\max_j q_j,
\end{equation}
where $q_j$ is recomputed from the current mask prediction at each iteration and detached during backpropagation. Presence-aware Confidence Supervision then consists of quality, ranking, and presence supervision:
\begin{align}
\mathcal{L}_{\mathrm{quality}} &= \frac{1}{N_q} \sum_{j=1}^{N_q} \mathcal{L}_{\mathrm{BCE}}(s_j,q_j),\\
\mathcal{L}_{\mathrm{ranking}} &= \max ( 0,\delta-s_{j^*}+\max_{j\neq j^*}s_j ),\\
\mathcal{L}_{\mathrm{presence}} &= \mathcal{L}_{\mathrm{BCE}}(s_p,1),
\end{align}
where $\delta$ is the ranking margin. Following the standard CD-FSS protocol, the target category is present in every reference and target image; therefore, the presence label is always positive. Further discussion is provided in the appendix. We combine all of these losses to optimize both the inverted reference tokens and the layer-wise fusion coefficients:
\begin{equation}
\mathcal{L}_{\mathrm{ref}} = \mathcal{L}_{\mathrm{mask}} +\lambda(\mathcal{L}_{\mathrm{quality}} +\mathcal{L}_{\mathrm{ranking}} +\mathcal{L}_{\mathrm{presence}}).
\end{equation}

Overall, Reference Inversion is optimized in two stages. We first learn $\mathbf{e}_w^*$ and $\mathbf{z}_c^*$ independently and then fix the resulting inverted reference tokens to optimize the layer-wise fusion coefficients. Throughout this process, SAM3 remains frozen and only approximately 12K category-specific parameters are learned. Once optimized, $\mathbf{T}_c^*$ is reused across target images, enabling single-pass inference without references or visual matching.
\section{Experiment}

\subsection{Dataset}

We conduct extensive experiments on four Cross-Domain Few-Shot Segmentation (CD-FSS) datasets spanning medical images~\cite{codella2019skin,tschandl2018ham10000,candemir2013lung,jaeger2013automatic}, satellite images~\cite{demir2018deepglobe}, and underwater scenes~\cite{islam2020semantic}. \textbf{ISIC2018}~\cite{codella2019skin,tschandl2018ham10000} consists of dermoscopic images for skin lesion analysis and covers three lesion categories. \textbf{Chest X-Ray}~\cite{candemir2013lung,jaeger2013automatic} is a dataset collected for tuberculosis screening. \textbf{DeepGlobe}~\cite{demir2018deepglobe} is a satellite image dataset containing six terrain categories, including urban, agriculture, rangeland, forest, water, and barren areas. Following PATNet~\cite{lei2022cross}, we split the original images into small patches for evaluation. \textbf{SUIM}~\cite{islam2020semantic} is an underwater dataset comprising six object categories, including fish, plants, divers, robots, ruins, and rocks.

\begin{table}[t]

\renewcommand\arraystretch{1.05}
\centering

\small
\setlength{\tabcolsep}{2.4pt}
\begin{tabular}{l||cccc}
\toprule[1pt]

Incorporated design & ISIC & Chest X-Ray & DeepGlobe & SUIM \\ \hline\hline
SAM3 $+$ category-name & 10.2 & 24.4 & 21.5 & 16.1 \\
SAM3 $+$ BRI & 54.9 & 89.3 & 23.8 & 54.4 \\
$+$ LTC & 59.8 & 89.5 & 43.9 & 58.8 \\
$+$ PCS & \textbf{64.1} & \textbf{90.5} & \textbf{51.8} & \textbf{63.2} \\
\bottomrule[1pt]

\end{tabular}

\caption{Ablation studies of proposed components of RI-SAM3 under the 1-shot setting (mIoU (\%)). All variants use the same pre-trained SAM3 with frozen parameters.}

\label{tab:ablation}
\end{table}

\begin{figure}[t]
    \centering
    \includegraphics[width=0.92\linewidth]{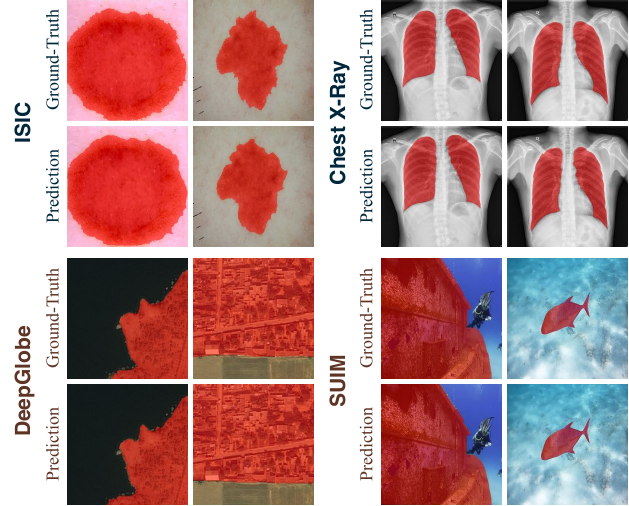}
    \caption{Qualitative segmentation results (red mask) of RI-SAM3 on samples from four datasets.}
    \label{fig:vis}
\end{figure}

\subsection{Implementation Details}

We adopt the official pre-trained SAM3 model~\cite{carion2025sam3} and keep it frozen. All images are processed at a resolution of $1008\times1008$. For each target category, the soft textual representation is learned solely from the annotated references. During inference, the reference image and mask are discarded, and each target image is processed with a single SAM3 forward pass conditioned on the learned textual representation. We report mean Intersection over Union (mIoU) as the evaluation metric. We strictly follow the standard evaluation protocols and dataset splits adopted by prior CD-FSS studies. RI-SAM3 only changes how the given references are utilized: instead of matching them with individual target images, it inverts the same reference set into a reusable textual condition.

\begin{table}[t]

\renewcommand\arraystretch{1.05}

\centering 
\small
\setlength{\tabcolsep}{2.6pt}

\begin{tabular}{l||cccc}
\toprule[1pt]

Method & ISIC & Chest X-Ray & DeepGlobe & SUIM \\\hline\hline
PerSAM-F & $20.5 \pm 6.3$ & $38.7 \pm 7.7$ & $46.5 \pm 10.5$ & $10.3 \pm 5.1$ \\
VLP-SAM & $34.8 \pm 5.0$ & $54.7 \pm 14.5$ & $48.1 \pm 16.1$ & $16.9 \pm 6.3$\\
\textbf{RI-SAM3} & $\textbf{64.1} \pm \textbf{4.8}$ & $\textbf{90.5} \pm \textbf{0.6}$ & $\textbf{51.8} \pm \textbf{2.9}$ & $\textbf{63.2} \pm \textbf{3.4}$\\

\bottomrule[1pt]
\end{tabular}

\caption{Stability comparison over five random reference selections, reported as mean mIoU (\%) $\pm$ standard deviation.}

\label{tab:stability}
\end{table}

\subsection{Comparison with State-of-the-Art Methods}

We compare our proposed RI-SAM3 with conventional ResNet-/ViT-based methods~\cite{fan2024adapting,su2024domain,lei2022cross,tong2024lightweight,herzog2024adapt,tong2025self,liu2025devil,nie2024cross,li2025dual} and Vision Foundation Model (VFM)-based methods~\cite{zhang2023personalize,liu2023matcher,he2024apseg,yang2024tavp,nie2026boosting,zhang2024bridge,cuttano2026insid3,sakurai2025vision}. As shown in Tab.~\ref{tab:cdfss}, RI-SAM3 achieves the best result in seven evaluation settings compared with VFM-based methods. On ISIC, our method improves the previous best VFM-based results by 9.2\% and 7.5\% under the 1-shot and 5-shot settings, respectively. Consistent gains are observed on Chest X-Ray, where it surpasses APSeg by 6.4\% and 6.5\%, and on SUIM, where it exceeds INSID3 by 8.3\% and 16.4\%. On DeepGlobe, RI-SAM3 achieves the best 1-shot result and remains competitive in the 5-shot setting. These results demonstrate the advantage of replacing visual matching with a representative textual condition. Our method also compares favorably with conventional ResNet-/ViT-based approaches. It reaches 90.5\% and 91.0\% mIoU on Chest X-Ray, outperforming the strongest results by 6.6\% and 3.9\%. Substantial improvements are also achieved on SUIM, demonstrating effectiveness under challenging underwater domain shifts. Moreover, RI-SAM3 obtains the best 5-shot result on ISIC and remains competitive on DeepGlobe. Finally, the 5-shot performance consistently exceeds its 1-shot counterpart across all datasets. This trend suggests that additional reference images provide more comprehensive information, enabling the learned textual condition more representative. As Fig.~\ref{fig:vis} shows, RI-SAM3 achieves accurate segmentation on CD-FSS datasets.

\subsection{Ablation Study}

All variants in Tab.~\ref{tab:ablation} use the same frozen SAM3 backbone, ensuring that their performance differences arise from the incorporated designs rather than backbone changes. Directly prompting SAM3 with category names yields limited performance under substantial domain shifts. Bi-level Reference Inversion (BRI) substantially improves ISIC, Chest X-Ray, and SUIM by 44.7\%, 64.9\%, and 38.3\%, respectively, demonstrating that the inverted textual condition is more representative than category names. Layer-adaptive Textual Conditioning further improves performance on ISIC, DeepGlobe, and SUIM by 4.9\%, 20.1\%, and 4.4\%, respectively, demonstrating the benefit of adaptively combining the inverted reference tokens across decoder layers. Finally, Presence-aware Confidence Supervision (PCS) consistently provides additional gains of 1.0\%-7.9\% by jointly supervising query confidence and target presence. Overall, the complete method consistently improves the SAM3 baseline prompted with category names by large margins across the four datasets, confirming the effectiveness and complementarity of the three designs. The appendix provides controlled comparisons with conventional textual inversion, visual prompt with SAM3, and GF-SAM using the frozen SAM3, fine-grained ablations of BRI and PCS, an analysis of shared versus layer-adaptive textual condition, and a non-empty prediction control. It also evaluates multiple textual initializations to verify that the improvements do not rely on carefully selected category labels.

\subsection{Robustness to Reference Image Selection}

In our RI-SAM3, the textual representations are inverted from the selected reference exemplar and subsequently reused across all target images. Robustness to reference image selection is therefore particularly important. We evaluate five random reference selections and compare with PerSAM-F~\cite{zhang2023personalize} and VLP-SAM~\cite{sakurai2025vision}, which respectively learn mask aggregation weights and multimodal reference prompts. As shown in Tab.~\ref{tab:stability}, RI-SAM3 achieves the highest mean performance and the lowest standard deviation across all datasets. The improvement is especially pronounced on Chest X-Ray and DeepGlobe, where our standard deviations are only 0.6 and 2.9. Consistent improvements are also observed on ISIC and SUIM. These results demonstrate that RI-SAM3 is substantially less sensitive to the choice of reference image, supporting reliable textual inversion across different reference selections. We also evaluate the robustness of our RI-SAM3 to different hyperparameters in the appendix.

\begin{table}[t]

\renewcommand\arraystretch{1.05}
\centering

\small

{\setlength{\tabcolsep}{14.2pt}
\begin{tabular}{l||ccc}
\toprule[1pt]
\multicolumn{4}{c}{\textit{\textbf{ResNet-based Methods}}}\\ \hline
\multirow{2}*{Method} & \multicolumn{3}{c}{Runtime (min)}\\ \cline{2-4}
& Preparation & Inference & Total\\ \hline\hline
PATNet & 538 & 5 & 543\\
IFA & 541 & 5 & 545\\\hline\hline

\end{tabular}}

{\setlength{\tabcolsep}{4.7pt}
\begin{tabular}{l||ccc|c|c}

\multicolumn{6}{c}{\textit{\textbf{SAM-based Methods}}}\\ \hline
\multirow{2}*{Method} & \multicolumn{3}{c|}{Runtime (min)} & \multirow{2}*{Model size (M)} & \multirow{2}*{FLOPs (T)}\\ \cline{2-4}
& Prep. & Inf. & Total & \\ \hline\hline
Matcher  & -- & 41 & 41 & 945 & 8.6\\
GF-SAM  & -- & 8 & 8 & 945 & 8.4\\
\textbf{RI-SAM3} & 2 & \textbf{2} & \textbf{4} & \textbf{840} & \textbf{3.7}\\

\bottomrule[1pt]
\end{tabular}}

\caption{Efficiency comparison under the 1-shot CD-FSS setting. Preparation refers to the optimization performed before target inference, including source training, target adaptation, or concept acquisition, depending on the method.}
  
\label{tab:efficiency}
\end{table}

\subsection{Efficiency Analysis}

Tab.~\ref{tab:efficiency} compares the efficiency of conventional and SAM-based CD-FSS methods. For a fair comparison, all efficiency evaluations are conducted on the same datasets under identical environments. Conventional methods, including PATNet and IFA, require over 500 minutes of preparation, resulting in total runtimes exceeding 540 minutes. In contrast, our method acquires a reusable textual condition in only 2 minutes while keeping SAM3 frozen and optimizing approximately 12K parameters. The detailed calculation of learnable parameters is provided in the appendix. Compared with SAM-based Matcher and GF-SAM, which require 41 and 8 minutes for inference, respectively, our method completes inference in only 2 minutes. Its total runtime is therefore only 4 minutes, while its model size is reduced from 945M to 840M parameters. Moreover, our method requires only $3.7\times10^{12}$ FLOPs, less than half the $8.6\times10^{12}$ and $8.4\times10^{12}$ FLOPs required by Matcher and GF-SAM. These results demonstrate that RI-SAM3 enables lightweight reference inversion and efficient text-conditioned inference.

\begin{table}[t]

\renewcommand\arraystretch{1.05}
\centering
\small
\setlength{\tabcolsep}{2.2pt}

\begin{tabular}{l||ccc|cc}
\toprule[1pt]
\multirow{2}*{Method} & \multicolumn{3}{c|}{In-Domain FSS} & \multicolumn{2}{c}{One-Shot Part Seg.} \\ \cline{2-6}
& Pasc.-5i & COCO-20i & FSS. & PASC.-Part & PACO-Part \\ \hline\hline
PerSAM & 43.1 & 23.0 & 71.2 & 32.5 & 22.5 \\
PerSAM-F & 48.5 & 23.5 & 75.6 & 32.9 & 22.7 \\
Matcher & 68.1 & 52.7 & 87.0 & 42.9 & 34.7 \\
VRP-SAM & 71.9 & 53.9 & -- & -- & -- \\
GF-SAM & 72.1 & 58.7 & 88.0 & 44.5 & 36.3 \\
CPS & -- & 49.2 & 81.7 & -- & -- \\
INSID3 & -- & 57.6 & -- & -- & -- \\
\textbf{RI-SAM3} & \textbf{80.4} & \textbf{70.9} & \textbf{91.5} & \textbf{57.6} & \textbf{45.8} \\

\bottomrule[1pt]
\end{tabular}

\caption{Generalization under the 1-shot setting in mIoU (\%). The best results are highlighted in \textbf{bold}.}
\label{tab:idfss}
\end{table}

\subsection{Generalization across Tasks}

Beyond CD-FSS, we evaluate the generalization of our proposed RI-SAM3 on three In-Domain Few-Shot Segmentation benchmarks and two One-Shot Part Segmentation benchmarks, detailed in the appendix. As shown in Tab.~\ref{tab:idfss}, our method consistently outperforms all Vision Foundation Model-based approaches across all settings. The complete comparison under both 1-shot and 5-shot settings is provided in the appendix. Compared with the strongest competing results, it improves PASCAL-5i by 8.3\% and 1.4\%, COCO-20i by 12.2\% and 7.6\%, and FSS-1000 by 3.5\% and 2.4\% under the 1-shot and 5-shot settings, respectively. It also surpasses GF-SAM by 13.1\% on PASCAL-Part and 9.5\% on PACO-Part. Qualitative examples in the appendix further illustrate accurate segmentation at both object and part levels. These consistent improvements across object- and part-level segmentation demonstrate that our RI-SAM3 learns category-level representations at different semantic granularities, indicating strong generalization beyond CD-FSS.

\section{Conclusion}

In this work, we revisited reference-target visual matching in CD-FSS and highlighted its sensitivity to target appearance variations. We proposed RI-SAM3, which replaces per-sample visual matching with reusable textual tokens inverted from annotated references. Bi-level Reference Inversion learns complementary tokens before and after the frozen text encoder, Layer-adaptive Textual Conditioning adapts their combination across decoder layers, and Presence-aware Confidence Supervision improves prediction reliability. With SAM3 frozen, RI-SAM3 optimizes only approximately 12K parameters per category and segments each target in a single forward pass without reference images. Extensive experiments demonstrate strong performance, robustness, computational efficiency, and generalization to related tasks.



\newpage

\bibliography{aaai2027}

\end{document}